\documentclass[letterpaper, 10 pt, conference]{ieeeconf}  

\IEEEoverridecommandlockouts                              

\usepackage{graphics} 
\usepackage{epsfig} 
\usepackage{mathptmx} 
\usepackage{times} 
\usepackage{amsmath} 
\usepackage{amssymb}  
\usepackage{siunitx}  
\usepackage{booktabs} 
\usepackage{multirow}
\usepackage{caption}
\usepackage{subcaption}

\usepackage[colorlinks = true,
        urlcolor  = blue]{hyperref}

 \usepackage{xurl}

\DeclareMathOperator*{\argmin}{arg\,min}

\title{\LARGE \bf
Decentralized Uncertainty-Aware Multi-Agent Collision Avoidance With Model Predictive Path Integral*}

\author{Stepan Dergachev$^1$ and Konstantin Yakovlev$^2$%
\thanks{*The reported study was supported by the Ministry of Science and Higher Education of the Russian Federation under Project 075-15-2024-544}
\thanks{$^1$Stepan Dergachev is with FRC CSC RAS and HSE University
{\tt\small dergachev@isa.ru}}%
\thanks{$^2$Konstantin Yakovlev is with FRC CSC RAS and AIRI 
{\tt\small yakovlev@isa.ru}}%
\thanks{This is a pre-print of the paper accepted to IROS2025.}%
}

\begin{document}

\maketitle
\thispagestyle{empty}
\pagestyle{empty}

\begin{abstract}

Decentralized multi-agent navigation under uncertainty is a complex task that arises in numerous robotic applications. It requires collision avoidance strategies that account for both kinematic constraints, sensing and action execution noise. In this paper, we propose a novel approach that integrates the Model Predictive Path Integral (MPPI) with a probabilistic adaptation of Optimal Reciprocal Collision Avoidance. Our method ensures safe and efficient multi-agent navigation by incorporating probabilistic safety constraints directly into the MPPI sampling process via a Second-Order Cone Programming formulation. This approach enables agents to operate independently using local noisy observations while maintaining safety guarantees. We validate our algorithm through extensive simulations with differential-drive robots and benchmark it against state-of-the-art methods, including ORCA-DD and B-UAVC. Results demonstrate that our approach outperforms them while achieving high success rates, even in densely populated environments. Additionally, validation in the Gazebo simulator confirms its practical applicability to robotic platforms. A source code is available at: \url{http://github.com/PathPlanning/MPPI-Collision-Avoidance}.

\end{abstract}

\section{INTRODUCTION}

Numerous real-world robotic applications involve multiple robots that simultaneously move in the environment to accomplish the mission. Often it is not possible to control such fleets of robots in a centralized manner, when a single controller issues commands to every robot based on the global knowledge of the system's state. In such cases each robot has to makes decisions on its own about what control commands to generate in order to exhibit cooperative behavior and, crucially, to avoid collisions with the other robots -- the problem known as multi-agent collision avoidance.

\begin{figure}[t]
    \centering
    \includegraphics[width=\columnwidth]{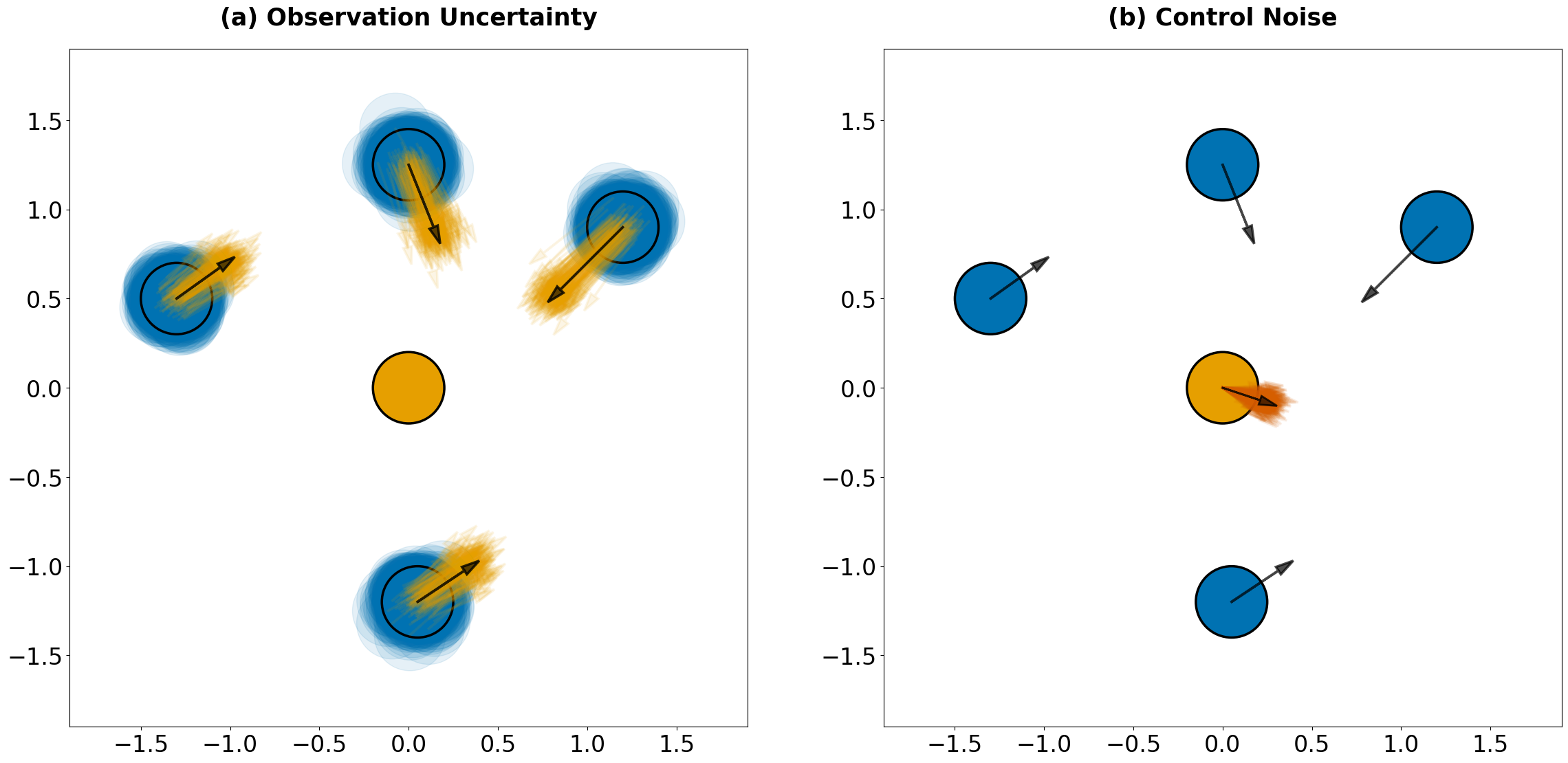}
    \caption{The decentralized multi-agent collision avoidance problem under uncertainty, where (a) illustrates observation uncertainty as any agent (orange in this case) perceives noisy estimates of others’ positions and velocities, while (b) depicts control noise, where execution errors perturbs the applied control.}
    \label{fig:abstract}
\end{figure}

Among the two major challenges associated with this problem is the need to respect the kinematic constraints of the robot and to handle uncertainty that is due to sensors' (Fig.~\ref{fig:abstract}-a) and actuators' noise (Fig.~\ref{fig:abstract}-b). Numerous multi-agent collision avoidance methods exist that rather ignore kinematic constraints~\cite{van2011reciprocal_n, zhou2017fast}, focus on specific dynamics~\cite{snape2011hybrid, van2011reciprocal}, or require extensive precomputation (e.g., lookup tables~\cite{alonso2013optimal}, additional controllers~\cite{zhu2022decentralized}, or fine-tuning~\cite{everett2021collision}). 

A prominent approach that is capable of taking kinematic constraints into account in the most general form (and thus is suitable for a wide range of robotic systems without extensive adaptation) is the Model Predictive Path Integral (MPPI)~\cite{williams2016aggressive}. MPPI relies on sampling-based optimization and originally is tailored to path following of a single robot~\cite{williams2017information}. Moreover in~\cite{dergachev2024model} a modification of MPPI was introduced that is not only tailored to multi-agent scenarios but is guaranteed to effectively provide safe outputs, i.e. such controls that respect the kinematic constraints and guarantee that the robot will not collide with the other robots. This is achieved by combining MPPI with ORCA~\cite{van2011reciprocal_n} a well-known collision-avoidance approach that guarantees safety. We dub this MPPI variant as MPPI-ORCA.

Unfortunately, MPPI-ORCA assumes perfect sensing (i.e. the ability to accurately estimate the positions and the velocities of the other robots) as well as perfect execution (i.e. it is assumed that the issued control commands are executed perfectly) -- the two assumptions that rarely met in the real-world robotics. As a result, when sensing and execution is not fully accurate MPPI-ORCA may lead to collisions. In this work we mitigate this problem.

Specifically, we present a collision avoidance method for a broad class of affine nonlinear systems that leverages MPPI and combines it with the safe sampling distribution adjustments. By ensuring sampled controls remain within a safe subset, our method improves sampling efficiency and safety. Moreover the arising optimization problem is a Second-Order Cone Programming that allows efficient solution by standard optimization solvers.

We extensively validate our approach in simulation. First, we use numeric simulation with the controlled injected noise to compare our method with state-of-the-art competitors such as ORCA for differential-drive robots~\cite{snape2010smooth}, B-UAVC~\cite{zhu2022decentralized} and MPPI-ORCA in a wide range of setups involving up to dozen of robots. Notably, our method significantly outperforms all the competitors. Then, we evaluate it in physical simulation, i.e. Gazebo, where we apply it to the well-known Turtlebot robots. We use Nav2 navigation stack and substitute its standard controller with ours. These experiments confirm that our method is able to provide safe controls and safely move the robots to their goals in the setups where the competitors fail.

We are committed to open-source our implementation as a ROS-node to make our results available to a wider robotics community.

\section{RELATED WORKS}

The following topics are especially relevant to our work: \textit{decentralized multi-agent navigation}, \textit{multi-agent collision avoidance}, \textit{model predictive path integral}.

\paragraph{Decentralized Multi-Agent Navigation} In decentralized approaches, agents make independent decisions based on local observations, communication, or a combination of both. Some methods operate under limited communication constraints and rely solely on observable information, such as agent positions~\cite{zhou2017fast} or both positions and velocities~\cite{van2011reciprocal_n}. Others incorporate additional information, such as planned trajectories or control actions of neighboring agents~\cite{zhu2019chance}. In decentralized settings, agents must independently select safe actions at each time step, necessitating the development of robust collision avoidance strategies.

\paragraph{Multi-Agent Collision Avoidance}

Velocity-based methods determine safe velocity regions and subsequently select the optimal velocity within these regions. The Optimal Reciprocal Collision Avoidance (ORCA) algorithm~\cite{van2011reciprocal_n} is a widely adopted approach for reciprocal collision avoidance. However, it does not account for kinematic constraints or uncertainty in agent positions. Works~\cite{snape2010smooth} and~\cite{snape2014smooth} describe the adaptation of the \textit{ORCA} algorithm to the differential-drive robot dynamics by enlarging the radius of the agent. The Non-Holonomic ORCA (NH-ORCA) algorithm~\cite{alonso2013optimal, alonso2018cooperative} extends ORCA to non-holonomic agents, such as differential-drive robots, but relies on precomputed lookup tables to incorporate motion constraints. The Probabilistic Reciprocal Velocity Obstacle (PRVO) method~\cite{gopalakrishnan2017prvo} addresses localization and observation uncertainty but does not consider kinematic constraints. The CALU~\cite{hennes2012multi} and COCALU~\cite{claes2012collision} methods integrate uncertainty-aware collision avoidance while also accommodating non-holonomic kinematics. However, these methods assume that agents exchange position and velocity data, which may not always be feasible in decentralized settings.

BVC-based methods~\cite{zhou2017fast} construct buffered Voronoi regions around agents to enforce safety. PBVC~\cite{wang2019distributed} extends this approach to a fully decentralized setting with observation uncertainty, but it does not account for motion constraints. B-UAVC~\cite{zhu2022decentralized} further improves BVC by incorporating positional uncertainty (assuming information exchange between agents) and can be adapted to specific dynamic models. However, in its general form, it relies on Model Predictive Control (MPC) with B-UAVC constraints, requiring the development of a suitable MPC controller tailored to the agent’s specific motion dynamics.

RL-based methods~\cite{long2017deep, long2018towards, chen2019crowd, fan2020distributed, everett2021collision} approximate value functions for safe and goal-directed motion. These methods typically operate directly on raw sensor data (e.g., LIDAR), eliminating the need for explicit velocity estimation. However, RL-based solutions often lack theoretical guarantees and require scenario-specific training.

\paragraph{Model Predictive Path Integral}
Originally introduced in~\cite{williams2016aggressive}, the Model Predictive Path Integral (MPPI) algorithm assumes a nonlinear affine dynamical system. Its development led to the Information-Theoretic Model Predictive Control (IT-MPC) algorithm~\cite{williams2017information}, which extends its applicability to general nonlinear systems. Subsequent research has focused on improving robustness and sampling efficiency~\cite{gandhi2021robust, balci2022constrained, tao2022control, tao2022path}.

For instance, \cite{tao2022control, tao2022path} enhance safety by incorporating Control Barrier Functions, while~\cite{kim2022smooth} explores methods for generating smoother trajectories. Another research direction addresses uncertainties in system dynamics, perception, and control execution~\cite{mohamed2025towards}. Some studies focus on dynamic collision avoidance, either assuming inter-agent communication~\cite{wang2022sampling, song2023safety, mohamed2025chance} or relying on motion prediction without formal safety guarantees~\cite{streichenberg2023multi}. The work presented in~\cite{dergachev2024model} tackles this problem by integrating collision avoidance constraints; however, it does not account for uncertainties in perception and execution errors in control actions.

\section{PROBLEM STATEMENT}
We follow~\cite{zhu2022decentralized} to formulate the problem. Specifically, let $\mathcal{A} = \{1, 2, \dots, N\}$ denote a set of homogenoeous robots (agents) operating in a workspace $\mathcal{W} \subset \mathbb{R}^2$ (please note, that our findings can be directly extended to 3D workspaces as well, but we concentrate on 2D for brevity). The safety radius of each robot is $r$. The time is discretized and at each time step a robot picks a control $u \in \mathbb{R}^m$ to change its state $\mathbf{x}_t \in \mathbb{R}^n$. The actual control signal, however, is assumed to be perturbed due to execution inaccuracies:

\begin{equation}
    \mathbf{\nu}_t \sim \mathcal{N}(\mathbf{u}_t, \Sigma), \quad \Sigma = \operatorname{diag}(\sigma_1^2, \dots, \sigma_m^2)
\end{equation}

Here $\nu_t \in \mathbb{R}^m$ is a randomly perturbed control that is actually executed and the motion of a robot is modeled as a discrete-time, continuous-state dynamical system of the following nonlinear affine form:

\begin{equation}
\label{eq:aff_dyn}
    \mathbf{x_{t+1}} = F(\mathbf{x_t}) + G(\mathbf{x_t})\mathbf{\nu_t},
\end{equation}

\noindent where $F : \mathbb{R}^n \to \mathbb{R}^n$ and $G : \mathbb{R}^n \to \mathbb{R}^{n \times m}$ are the given functions and $\nu$ is bounded: 

\begin{equation}
\label{eq:control_bounds}
\mathbf{\nu}_{\min}[k] \leq \mathbf{\nu}_t[k] \leq \mathbf{\nu}_{\max}[k], \quad k = 1, \dots, m.
\end{equation}

Here and throughout the work $[k]$ denotes the $k$-th component of a vector.

At each time step, each robot $i$ is assumed to have full knowledge of its own state $\mathbf{x}^i_t$, including its position $\mathbf{p_{t}^i}$. 

Robot $i$ can also sense the relative positions and velocities of the nearby robots, but this sensing is noisy. Specifically, the observed position $\hat{\mathbf{p}}^j_t$ and velocity $\hat{\mathbf{v}}^j_t$ of any neighboring robot $j$ are modeled as Gaussian distributions with covariances  $\Sigma_{p}$ and $\Sigma_{v}$, respectively, such that  $\hat{\mathbf{p}}^j_t \sim N(\mathbf{p}^j_t, \Sigma_{p})$ and $\hat{\mathbf{v}}^j_t \sim N(\mathbf{v}^j_t, \Sigma_{v})$.

Consider now a robot $i$ residing in state $\mathbf{x}^i_t$ and observing robot $j$: $(\hat{\mathbf{p}}^j_t, \hat{\mathbf{v}}^j_t)$. The control $\mathbf{u}^i_t$ is called \emph{probabilistically safe} (or simply \emph{safe}) w.r.t. robot $j$ if after executing $\mathbf{\nu}_t \sim \mathcal{N}(\mathbf{u}_t, \Sigma)$ and arriving at $\mathbf{x}^i_{t+1}$ the probability that the distance between $i$ and $j$ is less than $2 \cdot r$ does not exceed the predefined threshold $\delta$

\textbf{The problem} now is to consecutively determine a control input  $\mathbf{u}^i_t$  for each robot $i$ given its current state  $\mathbf{x}^i_t$  s.t. \textit{(i)} it satisfies the prescribed control constraints; \textit{(ii)} it ensures progress toward the given goal position $\tau_i$; \textit{(iii)} it is probabilistically safe with the respect to all observed robots.

\section{BACKGROUND}

As our method is based on combination of Model Predictive Path Integral (MPPI)~\cite{williams2016aggressive} and Optimal Reciprocal Collision Avoidance (ORCA) algorithm~\cite{van2011reciprocal_n} we first give a brief overview of these methods.

\subsection{Model Predictive Path Integral}

Let  $\mathcal{V} = (\mathbf{\nu_0}, \mathbf{\nu_1}, \ldots, \mathbf{\nu_{T-1}}) $ be the control sequence, $\mathcal{U}$ the set of admissible controls, and  $X = (\mathbf{x_0}, \mathbf{x_1}, \ldots, \mathbf{x_{T}})$  the state trajectory over horizon  $T$. The cost function is:
\begin{equation}
    \mathcal{L}(X, \mathcal{V}) = \phi(\mathbf{x_T}) + \sum_{t=0}^{T-1}\left(q(\mathbf{x_t}, \mathbf{u_t}) + \frac{\lambda}{2} \mathbf{\nu^T_t} \Sigma^{-1} \mathbf{\nu_t} \right)
\end{equation}

\noindent where  $\lambda \in \mathbb{R^+}$  is the inverse temperature parameter,  $\phi(\mathbf{x_T})$  the terminal cost, and  $q(\mathbf{x_t}, \mathbf{u_t})$  the running cost. Each  $\mathbf{\nu_t} \sim \mathcal{N}(\mathbf{u_t}, \Sigma)$, with $\mathbf{u_t}$ is control input and $\Sigma$  as the system’s noise covariance. The optimal control problem minimizes the expected cost:
\begin{equation}
    U^* = \argmin_{\mathcal{V} \in \mathcal{U}} \mathbb{E}_\mathbb{Q}[\mathcal{L}(X, \mathcal{V})]
\end{equation}

Given the current state  $\mathbf{x_0}$  and an initial control sequence  $U^{init} = (\mathbf{u^{init}_0}, \ldots, \mathbf{u^{init}_{T-1}})$. The MPPI algorithm samples  $K$  noise sequences  $\xi^k = (\mathbf{\epsilon^k_0}, \ldots, \mathbf{\epsilon^k_{T-1}}),\; \mathbf{\epsilon^k_t} \sim \mathcal{N}(0, \Sigma^*)$. Notably that the sampling variance $ \Sigma^*$ may be higher than the natural variance $\Sigma$ (but also should be diagonal). Based on $U^{init}$ and $\xi^k$, a set of $K$ control sequences $U^k$ is obtained: 
\begin{equation}
U^k = (\mathbf{u^k_0}, \ldots, \mathbf{u^k_{T-1}}), \quad \mathbf{u^k_t} = \mathbf{u^{init}_t} + \mathbf{\epsilon^k_t}
\end{equation}

Each  $U^k$  generates a trajectory  $X^k$, evaluated by $\tilde{S}(U)$:
\begin{equation}
\begin{aligned}
S(U) = & \phi(\mathbf{x_T}) + \sum_{t=0}^{T-1} q(\mathbf{x_t}, \mathbf{u_t})\\
\tilde{S}(U) = & S(U) + \frac{\lambda}{2} \sum_{t = 0}^{T-1} \left( \right.\mathbf{u^T_t} \Sigma^{-1} \mathbf{u_t} + \\
& 2\mathbf{u^T_t} \Sigma^{-1} \mathbf{\epsilon_t} + \mathbf{\epsilon^T_t}(I-\mathcal{K}^{-1})\Sigma^{-1}\mathbf{\epsilon_t} \left.\right)
\end{aligned}
\end{equation}
\noindent where $I$ is identity matrix and $\mathcal{K}$ is a multiplier on the magnitude of the sampling variance ($\Sigma^*=\mathcal{K}\Sigma$).

Weights for each trajectory are computed as:
\begin{equation}
\omega(U) = \frac{\exp\left(-\frac{1}{\lambda} (\tilde{S}(U) - \rho)\right)}{\sum_{k=1}^{K} \exp\left(-\frac{1}{\lambda} (\tilde{S}(U^k) - \rho) \right)}
\end{equation}
\noindent where $\rho = \min_{k}{S(U^k)}$. The final control sequence is:
\begin{equation}
\mathbf{u^{*}_t} = \sum_{k = 1}^{K} \omega(U^k) \mathbf{u^k_t}
\end{equation}

After execution of  $\mathbf{u^*_0}$,  $U^{init}$  is updated:
\begin{equation}
U^{init} = (\mathbf{u^*_1}, ..., \mathbf{u^*_{T-1}}, \mathbf{u^{init}})
\end{equation}

MPPI efficiently solves nonlinear stochastic control problems but may yield unsafe solutions in multi-agent settings due to trajectory collisions, requiring more samples for safety.

\subsection{Optimal Reciprocal Collision Avoidance}

The Optimal Reciprocal Collision Avoidance (ORCA) method~\cite{van2011reciprocal_n} extends the velocity obstacles theory~\cite{fiorini1998motion} to enable decentralized, reciprocal multi-agent collision avoidance. ORCA computes collision-free velocities based on observable information, assuming all agents follow the same policy.

The algorithm performs safe movement in the following way. The agent updates its state/observation and constructs a set of feasible velocities bounded by linear constraints, and uses linear programming to find the closest velocity to a preferred one. The chosen velocity is then executed, and the state is updated. Let's look at the basic ORCA approach in more detail.

For two agents $i, j$, linear constraints are derived as follows. Let $\mathbf{v^i_t}$ denote the agent’s velocity $\mathbf{v^i_t} = ({\mathbf{p_{t+1}^i} - \mathbf{p_t^i}})$.

The velocity obstacle  $VO^{\tau}_{i|j}$  defines relative velocities leading to a collision within time  $\tau$:
\begin{multline}
VO^{\tau}_{i|j} = \{\mathbf{v} \; | \; \exists t' \in [0, \tau] : t'\cdot \mathbf{v} \in D(\mathbf{p_t^j} - \mathbf{p_t^i}, r^i + r^j)\}
\end{multline}
\noindent where $D(\mathbf{p}, r)$ is a disk with center position $p$ and radius $r$.

On the basis of the $VO^{\tau}_{i|j}$ a linear constraint $ORCA^{\tau}_{i|j}$ is constructed as follows. Assume that some optimization velocities $\mathbf{v^i_{opt}}, \mathbf{v^j_{opt}}$ will lead the agents to a collision at time $t' : (t' - t) < \tau$ (i.e. $\mathbf{v^{rel}_{i|j}} = (\mathbf{v^i_{opt}} - \mathbf{v^j_{opt}}) \in VO^{\tau}_{i|j}$). Let $\mathbf{u}$ be a vector to the nearest point on the boundary of $VO^{\tau}_{i|j}$ from $\mathbf{v_{rel}^{i|j}}$. In other words, $\mathbf{u}$ can be interpreted as the smallest change in $\mathbf{v_{rel}^{i,j}}$ to prevent a collision within time $\tau$.
\begin{equation}
\mathbf{u} = (\argmin_{\mathbf{v} \in \partial VO^{\tau}_{i|j}} || \mathbf{v} - \mathbf{v_{rel}^{i|j}}||) - \mathbf{v_{rel}^{i|j}}
\end{equation}

Next, the set of collision-free velocities is then defined as:
\begin{equation}
ORCA^{\tau}_{i|j} = \{ \mathbf{v} \; | \; (\mathbf{v} - (\mathbf{v^i_{opt}} + \alpha_{resp}, \mathbf{u})) \cdot \mathbf{n} \geq 0\}
\end{equation}

\noindent where $\mathbf{n}$ is the normal vector at the point nearest to $\mathbf{v_{rel}^{i|j}}$ at the boundary of $VO^{\tau}_{i|j}$, $\alpha_{resp}$  is the responsibility factor, set to  $\geq 1/2$  for reciprocal avoidance and  $1$  for static obstacles.

By enforcing symmetrical linear constraints in velocity space, ORCA ensures agents avoid collisions efficiently while reducing deadlocks. However, the method is limited to velocity-space solutions and may not account for kinematic constraints or sensing uncertainty.

\section{METHOD}

\begin{figure}[t]
    \centering
    \includegraphics[width=\columnwidth]{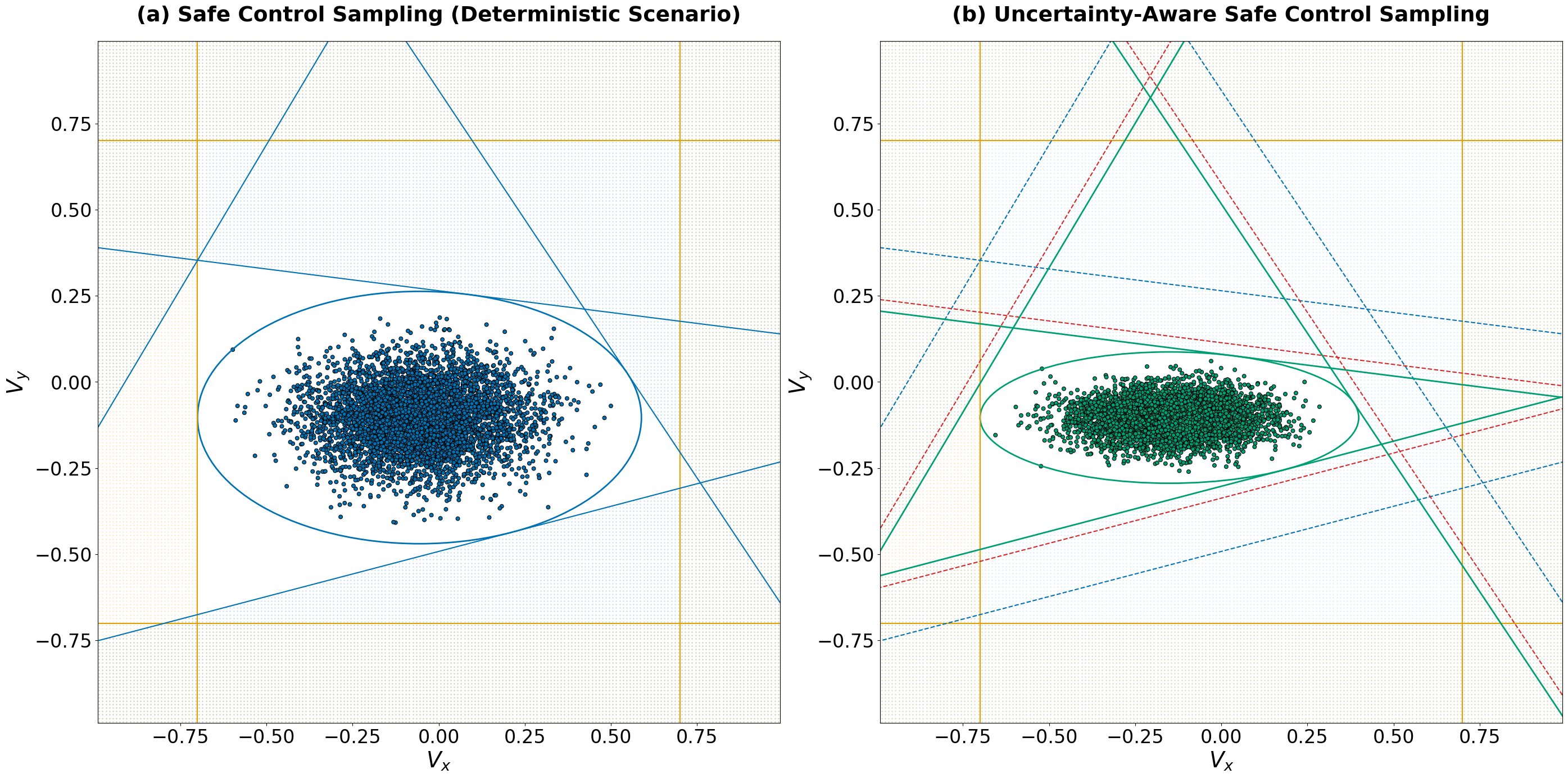}
    \caption{An example of safe control sampling (velocity sampling for single-integrator dynamics). (a) The deterministic scenario, as discussed in~\cite{dergachev2024model}, where control samples are generated within safety linear constraints (blue lines) and control bounds (orange lines). (b) The uncertainty-aware scenario, where additional buffers are incorporated to account for inaccuracies in observations and control execution. Specifically, the deterministic constraints (blue dashed lines) are expanded with buffers that account for observation errors (between blue dashed lines and red dashed lines) and control execution errors (between red dashed lines and green lines), resulting in a more conservative safe control region.}
    \label{fig:method}
\end{figure}

Our method for multi-robot safe collision avoidance under execution and observation uncertainty is based on combining ORCA linear constraints (i.e. providing safety) with MPPI, similarly to~\cite{dergachev2024model}. Specifically, we extend~\cite{dergachev2024model} (Fig.~\ref{fig:method}-a) to explicitly account for both observation inaccuracies and execution errors in selected control actions. Our method introduces additional buffers to the linear constraints, with buffer sizes computed based on the known distributions of errors in observations and control execution (Fig.~\ref{fig:method}-b).

We first enhance the ORCA constraints to account for uncertainty in the estimated states of other agents. Subsequently, we provide a detailed description of the integration of linear constraints into the MPPI sampling process, including the incorporation of control execution errors.

\subsection{ORCA Constraints with Respect to Observation Uncertainty}
\label{subsec:orca_unc}

The original ORCA algorithm assumes perfect sensing and does not account for perception uncertainty, but this can be mitigated with modifications.

First, uncertainty in velocity measurements can be managed by adjusting the optimization velocity  $v^i_{opt}$. While the agent’s current velocity  $v^i$  is typically used, setting  $v^i_{opt} = 0$  ensures collision avoidance when reliable velocity data is unavailable.

Second, sensing uncertainty can be addressed by inflating the agent’s radius with an uncertainty buffer  $r_{unc}$, resulting in an adjusted radius  $\hat{r}^i = r^i + r_{unc}$. The buffer size should be determined based on observation noise to maintain a sufficient safety margin.

When selecting a velocity that satisfies the ORCA linear constraints, we ensure that it remains outside the velocity obstacle  $VO^{\tau}_{i|j}$, which is constructed based on the agent’s position  $\mathbf{p}^i$, the estimated position  $\hat{\mathbf{p}}^j$  of the other agent, and their respective radii  $r^i$  and  $r^j$.

Let  $\hat{\mathbf{p}}^j$  be a realization of a random vector with mean  $\mathbf{p}^j$  and covariance  $\Sigma_p$ . Then, with probability at least $\delta_o$, the true position  $\mathbf{p}^j$  lies within the disk  $D(\hat{\mathbf{p}}^j, r_o)$, where the radius $r_o$ is given by:
\begin{equation}
    r_o = \sqrt{\lambda_{\max} {\chi^2_2}^{-1}(\delta_{o})},   
\end{equation}

\noindent where $\lambda_{\max}$ is the largest eigenvalue of $\Sigma_p$, ${\chi^2_2}^{-1}(\delta_{o})$ is the chi-square quantile (with 2 degrees of freedom) corresponding to probability $\delta_o$. 

To account for this uncertainty, we define an augmented velocity obstacle  $\widehat{VO}^{\tau}_{i|j}$ using the adjusted radius $\hat{r}^i = r^i + r_o$

It is straightforward to see that all possible velocity obstacles corresponding to any realization of  $\mathbf{p}^j$  within  $D(\hat{\mathbf{p}}^j, r_o)$ lie inside  $\widehat{VO}^{\tau}{i|j}$. Thus, by setting  $r_{unc} = r_o$, we can construct an ORCA linear constraint that guarantees collision avoidance with at least a probability $\delta_o$ .

\subsection{Incorporation Safety Constraints into the MPPI Sampling Framework}

To enforce collision avoidance, we adjust the sampling distribution parameters $\mathbf{\mu}, \Sigma^*$ to stay close to the original while ensuring a high probability of constraint satisfaction. This is formulated as an optimization problem.

\paragraph{Objective Function}
First, we need to define the closeness of the new distribution parameters $\mathbf{\mu'}, \Sigma'$ for $\mathbf{u_{t}}$ to the initial parameters $\mu, \Sigma^*$. In this paper, we assume that the components of $\mathbf{u_{t}}$ are independently normally distributed. Thus, the closeness can be measured using the 1-norm or Chebyshev norm:
\begin{equation}
    || \mathbf{\mu'} - \mathbf{\mu}|| + || diag(\Sigma') - diag(\Sigma^*)||
\end{equation}

Additionally, since the elements of $diag(\Sigma')$ must be non-negative, we introduce the following constraints:

\begin{equation}
    \Sigma'_{k, k} \geq 0, \; k = 1,...m
    \label{eq:sigma_pos}
\end{equation}

\paragraph{Collision Avoidance Constraints}
Let $\mathcal{A}'_i$ represent the set of visible neighbors of an agent $i$. We denote a single linear constraint $ORCA^\tau_{i|j}$ for agent $i$ relative to another agent $j \in \mathcal{A}'_i$ as a triplet $(a_j, b_j, c_j)$. The velocity vector $(v^i_{x,t}, v^i_{y, t})$ satisfies this constraint if:
\begin{equation}
    a_j \cdot v^i_{x, t} + b_j \cdot v^i_{y, t} + c_j \leq 0
    \label{eq:constraint_inequality}
\end{equation}

Since the state $\mathbf{x_t}$ includes the agent's position $(p_{x, t}, p_{y, t})$, the velocity vector $(v^i_{x,t}, v^i_{y, t})$ can be rewritten to depend linearly on the control signal $\mathbf{\nu^i_t} = \mathbf{u}^i_t + \mathbf{\epsilon}^i_t$, where $\mathbf{\epsilon}^i_t \sim \mathcal{N}(0, \Sigma)$. For a fixed state $\mathbf{x^i_t}$, the terms $F^i(\mathbf{x^i_t})$ and $G^i(\mathbf{x^i_t})$ can be treated as a constant vector $\mathbf{F^i}$ and matrix $G^i$:
\begin{equation}
    \begin{matrix}
    v^i_{x, t} = \mathbf{F^i}[1] + G^i_{1, 1}\,\mathbf{\nu}^i_{t}[1] + ... + G^i_{1, m} \; \mathbf{\nu}^i_{t}[m] - p_{x, t}\\
    v^i_{y, t} = \mathbf{F^i}[2] + G^i_{2, 1}\,\mathbf{\nu}^i_{t}[1] + ... + G^i_{2, m} \; \mathbf{\nu}^i_{t}[m] - p_{y, t}\\
    \end{matrix} 
    \label{eq:velocity_control}
\end{equation}

Defining $\mathbf{a'_j}, b'_j$ as
\begin{align*}
    &\mathbf{a'_j} = \{ (a_j G_{1, 1} + b_j G_{2, 1}), ..., (a_j G_{1, m} + b_j G_{2, m})\} \\
    &b'_j = -(c_j + a_j (\mathbf{F^i}[1] - p_{x, t}) + b_j (\mathbf{F^i}[2] - p_{y, t}))
\end{align*}
\noindent we can rewrite the collision avoidance constraint:
\begin{equation}
     \mathbf{a'_j}^T \, \mathbf{u^i_{t}} \leq b'_j - \mathbf{a'_j}^T \, \mathbf{\epsilon}^i_t
    \label{eq:constraint_inequality_control}
\end{equation}

\paragraph{Chance Constraints Formulation}

Next, we transform the inequality~(\ref{eq:constraint_inequality_control}) into a chance constraint with the required confidence level $\delta_{\nu}$:
\begin{equation}
     Pr(\mathbf{a'_j}^T \, \mathbf{u^i_{t}} \leq b'_j - \mathbf{a'_j}^T \, \mathbf{\epsilon}^i_t) \geq \delta_{\nu}
\label{eq:constraint_inequality_control_pr}
\end{equation}
\noindent which, in turn, can be converted into a linear constraint:
\begin{equation}
     \mathbf{a'_j}^T \, \mathbf{u^i_{t}} \leq b'_j - \Phi^{-1}(\delta_{\nu})\sqrt{\mathbf{a'_j}^T \Sigma \mathbf{a'_j}}     
\label{eq:constraint_inequality_control_pr_lin}
\end{equation}
\noindent where $\Phi(\cdot)$ is the standard normal cumulative distribution function.

By considering $\mathbf{u^i_{t}}$ as a stochastic vector with independently normally distributed components, the chance constraint $Pr(\mathbf{a'_j}^T \,\mathbf{u^i_{t}} \leq b'_j - \Phi^{-1}(\delta_{\nu})\sqrt{\mathbf{a'_j}^T \Sigma \mathbf{a'_j}}) \geq \delta_{u}$ holds \textit{if and only if}:
\begin{equation}
    \mathbf{a'_j}^T \, \mathbf{\mu'} + \Phi^{-1}(\delta_{u}) \sqrt{\mathbf{a'_j}^T \Sigma' \mathbf{a'_j}} \leq b'_j - \Phi^{-1}(\delta_{\nu})\sqrt{\mathbf{a'_j}^T \Sigma \mathbf{a'_j}},
    \label{eq:orca_chance_constraint}
\end{equation}

We also incorporate control limits as chance constraints:
\begin{equation}
    \begin{array}{l}
    \mathbf{\mu'}[k] + \Phi^{-1}(\delta_{u}) \sqrt{\Sigma'_{k, k}} \; \leq \; u[k]_{max}, \\
    \mathbf{\mu'}[k] - \Phi^{-1}(\delta_{u}) \sqrt{\Sigma'_{k, k}} \; \geq \; u[k]_{min}, \\
    k = 1,...,m,
    \end{array}
    \label{eq:control_bound}
\end{equation}

Thus, the optimization problem for determining the new distribution parameters for agent $i$ can be formulated as follows:
\begin{equation}
    \begin{aligned}
        \argmin_{\mathbf{\mu'}, \Sigma'} \quad & || \mathbf{\mu'} - \mathbf{\mu}|| + || diag(\Sigma') - diag(\Sigma^*)||\\
        \textrm{s.t.} \quad & \mathbf{a'_j}^T \, \mathbf{\mu'} + \Phi^{-1}(\delta_{u}) \sqrt{\mathbf{a'_j} \Sigma' \mathbf{a'^T_j}} \; \leq \; b'_j \; ... \\ &  - \Phi^{-1}(\delta_{\nu})\sqrt{\mathbf{a'_j}^T \Sigma \mathbf{a'_j}},\; \forall j\in \mathcal{A'}_i \\
        \quad & \mathbf{\mu'}[k] + \Phi^{-1}(\delta_{u}) \sqrt{\Sigma'_{k,k}} \; \leq \; u_{max}[k], \; k = 1,...,m \\
    \quad & \mathbf{\mu'}[k] - \Phi^{-1}(\delta_{u}) \sqrt{\Sigma'_{k,k}} \; \geq \; u_{min}[k], \; k = 1,...,m \\
        \quad & \Sigma'_{k,k} \; \geq 0 \;, \; k = 1,...,m \\
    \end{aligned}
    \label{eq:opt_problem}
\end{equation}

By solving this optimization problem, we obtain new parameters for the sampling distribution, ensuring that with probability at least $\delta_c = \delta_o \times \delta_{\nu} \times \delta_u$, the sampled controls will be safe.

The optimization problem formulated above is convex and can be expressed as a Second-Order Cone Programming (SOCP) problem, following an approach similar to that described in~\cite{dergachev2024model}. Additionally, for certain types of dynamic systems, such as differential-drive robots, it can be reformulated as a Linear Programming (LP) problem.

\subsection{Cost Function Design}

In MPPI-based algorithms, the terminal cost $\phi(\cdot)$ and the state-dependent running cost $q(\cdot)$ play a crucial role in guiding the robot’s behavior by penalizing undesirable states. To ensure safe navigation and effective collision avoidance, each sampled trajectory is evaluated using the following approach:
\begin{equation}
    \phi(\mathbf{x_t}) = w_{term} \cdot q_{goal}(\mathbf{x_t})
\end{equation}
\begin{multline}
     q(\mathbf{x_t}, \mathbf{u_t})   = w_{goal} \cdot q_{goal}(\mathbf{x_t}) + w_{dist} \cdot q_{dist}(\mathbf{x_t}) + \\ w_{col} \cdot q_{col}(\mathbf{x_t}) + w_{vel} \cdot q_{vel}(\mathbf{u_t})
\end{multline}

\noindent where $q_{goal}$ encourages the agent to progress toward its goal, $q_{dist}$ and $q_{col}$ impose penalties for potential collisions and $q_{vel}$ penalizes low speeds to promote efficient movement. $w_{term}, w_{goal}, w_{dist}, w_{col}, w_{vel}$ are the weighting factors that controls the influence of each component in the cost function.

Notably, the  $q_{dist} ,  q_{col}$, and  $q_{vel}$  terms are disabled when the agent is close to its target position, ensuring that unnecessary penalties do not interfere with goal-reaching behavior.

\paragraph{Toward Goal Progress Cost}

The goal progress cost quantifies how effectively the agent moves toward its target $\tau$. It is defined based on the Euclidean distance between the agent’s  position $\mathbf{p_t}$  and a projected goal position $\mathbf{\tau^{pr}}$:
\begin{equation}
\begin{matrix}
    q_{goal}(\mathbf{x_t}) = ||\mathbf{p_t} - \mathbf{\tau^{pr}}||_2
\end{matrix}
\end{equation}

\noindent $\mathbf{\tau^{pr}}$ represents the projection of the target onto a circle of diameter $d_{la}$, centered at the initial position $\mathbf{x_0}$ and $d_{la}$ is the look-ahead distance, which defines how far ahead the agent should consider its goal projection.

\paragraph{Distance to Agents Cost}

The distance to agents cost helps prevent potential collisions by encouraging agents to maintain a safe separation from one another. This cost function relies on neighboring agents’ predicted positions, which are estimated using a constant velocity model and a Kalman filter.

The cost penalizes states where the agent is too close to predicted neighbors:
\begin{equation}
q_{dist}(\mathbf{x_t}) =
\begin{cases}
    0,\; \text{if} \; ||\mathbf{x_t} - \mathbf{x^j_t}||_2 > d_{th} \\
    1 / \min_{j\in \mathcal{A'}}(||\mathbf{x_t} - \mathbf{\hat{x}^j_t}||^2_2), \; \text{otherwise}
    \end{cases}
\end{equation}
    
\noindent where $\mathcal{A'}$ is the set of neighboring agents, $\mathbf{\hat{x}^j_t}$ represents the predicted position of another agent and $d_{th}$ is the safety threshold distance.

\paragraph{Collision Cost}

The collision cost applies a strong penalty when a collision is predicted, ensuring that the agent avoids dangerous states. Unlike  $q_{dist}(\mathbf{x_t})$, which softly penalizes proximity, this cost strictly enforces safety when a collision is imminent.

Unlike deterministic models, collision estimation must account for uncertainty in agent positions. A popular approach is using Mahalanobis distance to approximate collision probability~\cite{toit2011probabilistic, mohamed2025chance}. However, for agents with a total size  $2r > 0.3\,m$, this method underestimates the probability of collisions.

To mitigate this issue, we employ an estimation method based on Euclidean distance with an uncertainty buffer  $r_{unc}$. The buffer $r_{unc}$ (similar to one described in Section~\ref{subsec:orca_unc}) is derived from the variance predicted by the Kalman filter, ensuring that position uncertainty is considered.
\begin{equation}
q_{col}(\mathbf{x_t}) =
\begin{cases}
1, & \text{if } ||\mathbf{x_t} - \mathbf{\hat{x}^j_t}||_2 < 2r + r_{unc}\\
0, & \text{otherwise}
\end{cases}
\end{equation}

\paragraph{Velocity Cost}

The velocity cost prevents the agent from moving too slowly. This cost function applies a penalty inversely proportional to the velocity magnitude:
\begin{equation}
q_{vel}(\mathbf{u_t}) = \frac{1}{||\mathbf{u_t}||_2}
\end{equation}

\section{EXPERIMENTAL EVALUATION}

The proposed method was implemented in C++ and evaluated in various experimental setups. Initially, experiments were conducted using a widely adopted model of differential-drive robot motion. This system model has been extensively studied in the literature on collision avoidance with kinematic constraints~\cite{snape2011hybrid, snape2010smooth, zhu2022decentralized}, enabling a comparative analysis with several well-established methods. Subsequently, the algorithm was integrated into the ROS2 Navigation stack as a controller module and validated using a well-known Gazebo robotic simulator.

\subsection{Comparison with Other Collision Avoidance Algorithms}

\subsubsection{Experimental Setup}

In the first series of experiments, a differential-drive robot model was utilized. The agent’s state, denoted as $\mathbf{x}$, was defined as $\mathbf{x} = (p_x, p_y, \theta)^T$, where $p_x, p_y$ represent the position of the robot (center of the corresponding disk) in a 2D workspace, and $\theta$ denotes the robot’s heading angle. The control input was given by $\mathbf{u} = (v, w)^T$, where $v$ is the linear velocity and $w$ is the angular velocity. At each time step, the selected control input was perturbed by normal noise $\varepsilon \sim \mathcal{N}(0, \Sigma)$. The resulting executed control was subject to bounding constraints analogous to those in~(\ref{eq:control_bounds}):

\begin{equation}
v_{min} \leq (v + \varepsilon[0]) \leq v_{max}, \quad w_{min} \leq (w + \varepsilon[1]) \leq w_{max}.
\end{equation}

The robot’s motion was governed by the following equations:

\begin{equation}
\mathbf{x}_{t + 1} = \mathbf{x}_t +
\begin{pmatrix}
\cos{\theta_t} & 0\\
\sin{\theta_t} & 0\\
0  & 1
\end{pmatrix} (\mathbf{u_t} + \varepsilon).
\end{equation}

For all robots in the experiments, the following parameters were used: robot size (radius of the corresponding disk) $r = 0.3 m$, linear velocity limits $v_{min} = -1.0\,m/s, v_{max} = 1.0\,m/s$, angular velocity limits $w_{min} = -2.0\,rad/s,\;  w_{max} = 2.0 \, rad/s$, and control noise covariance $\Sigma = \text{diag}(0.1 \,{m/s}, 0.2\,{rad/s})^2$. The observation radius was not restricted, but the observed positions and velocities of the neighboring agents were supplemented with noise with zero mean and covariance $\Sigma_p = \Sigma_v = \text{diag}(0.1, 0.1)^2$.

Two types of scenarios, widely used in literature on collision avoidance, were considered: \texttt{Circle} and \texttt{Random}. In \texttt{Circle} scenario 2-15 agents were positioned equidistantly along the circumference of a circle with a diameter of $D_{\text{circle}} = 12 m$. The objective of each agent was to navigate to the diametrically opposite position within the circle. The initial heading angle of each agent was aligned toward its respective goal position. In \texttt{Random} scenario 5-25 robots were randomly placed inside a $20\,m \times 20\,m$ area and their target locations were assigned randomly. 10 different instances were generated per each number of agents.

An instance was considered to be solved successfully if every agent safely (i.e. without collisions) reached its goal with tolerance $0.4\,m$ and the number of simulation steps did not exceed a predefined limit  of $1000$ steps. The duration of each simulation time step was fixed at $0.1\,s$. Each instance was executed $10$ times to ensure statistical significance of the results.

\begin{table}[t]
\centering
\begin{tabular}{l|cc|cc}
\toprule
 \multirow{2}{*}{\textbf{Algorithm}} & \multicolumn{2}{c|}{\texttt{Circle}} & \multicolumn{2}{c}{\texttt{Random}} \\
  & \textbf{SR} & \textbf{\% Col.} & \textbf{SR} & \textbf{\% Col.}\\

\midrule
ORCA-DD & 74.3\% & \textbf{0\%} & 98\% & \textbf{0\%}\\
B-UAVC &  99.3\% & 0.7\% & 85.4\% & \textbf{0\%}\\
MPPI-ORCA & 85\% & 15\% & 96.8\% & 3.2\%\\
\midrule
Ours & \textbf{100\%} & \textbf{0\%} & \textbf{100\%} & \textbf{0\%}\\
\bottomrule
\end{tabular}
\caption{\textbf{Success rate} and the \textbf{percentage of runs that failed due to collisions} of the evaluated algorithms.}
\label{tab:sr_comparison}
\end{table}

\begin{figure}[t]
     \centering
     \begin{subfigure}[b]{0.49\columnwidth}
         \centering
    \includegraphics[width=\textwidth]{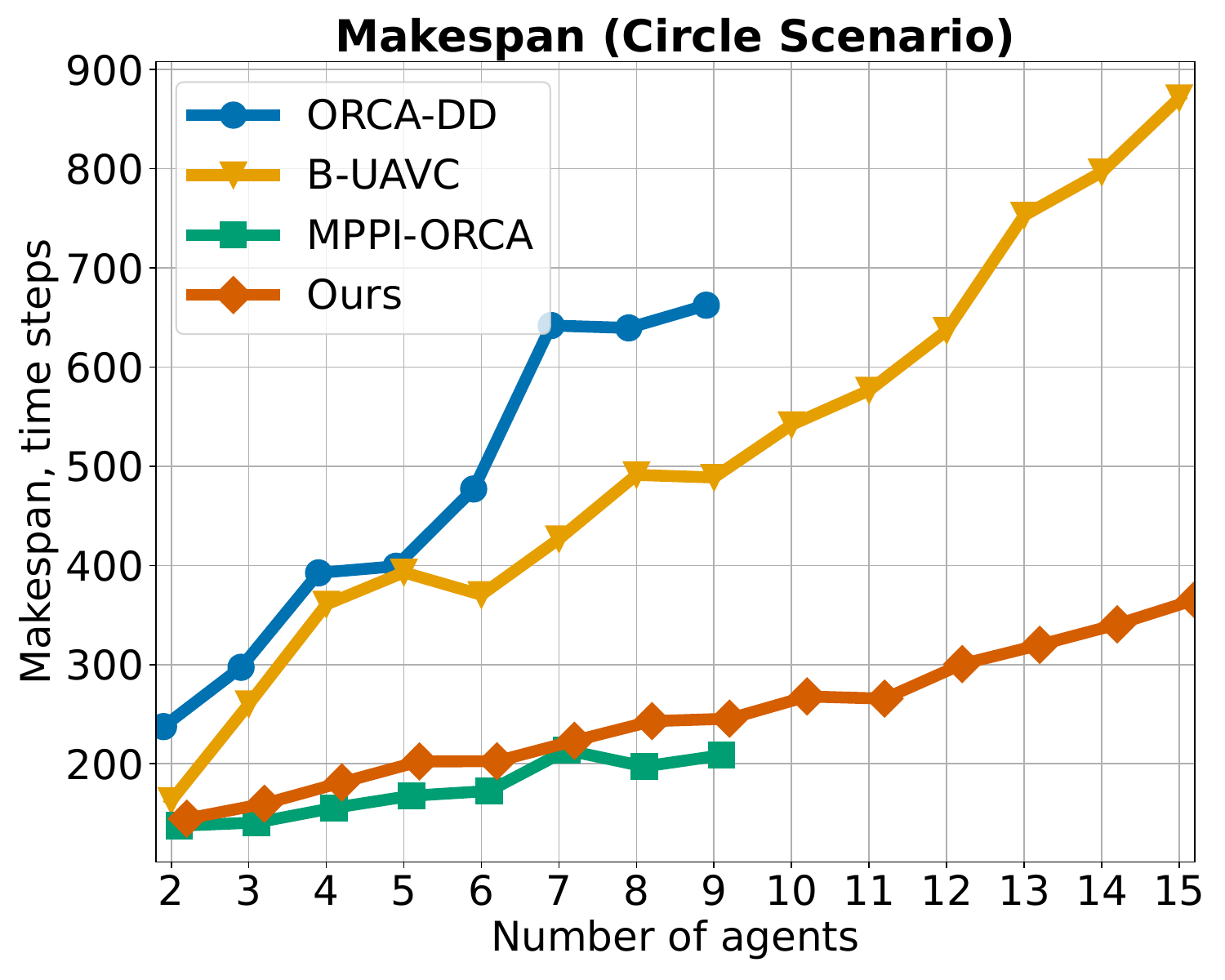}
     \end{subfigure}
     \hfill
     \begin{subfigure}[b]{0.49\columnwidth}
         \centering
         \includegraphics[width=\textwidth]{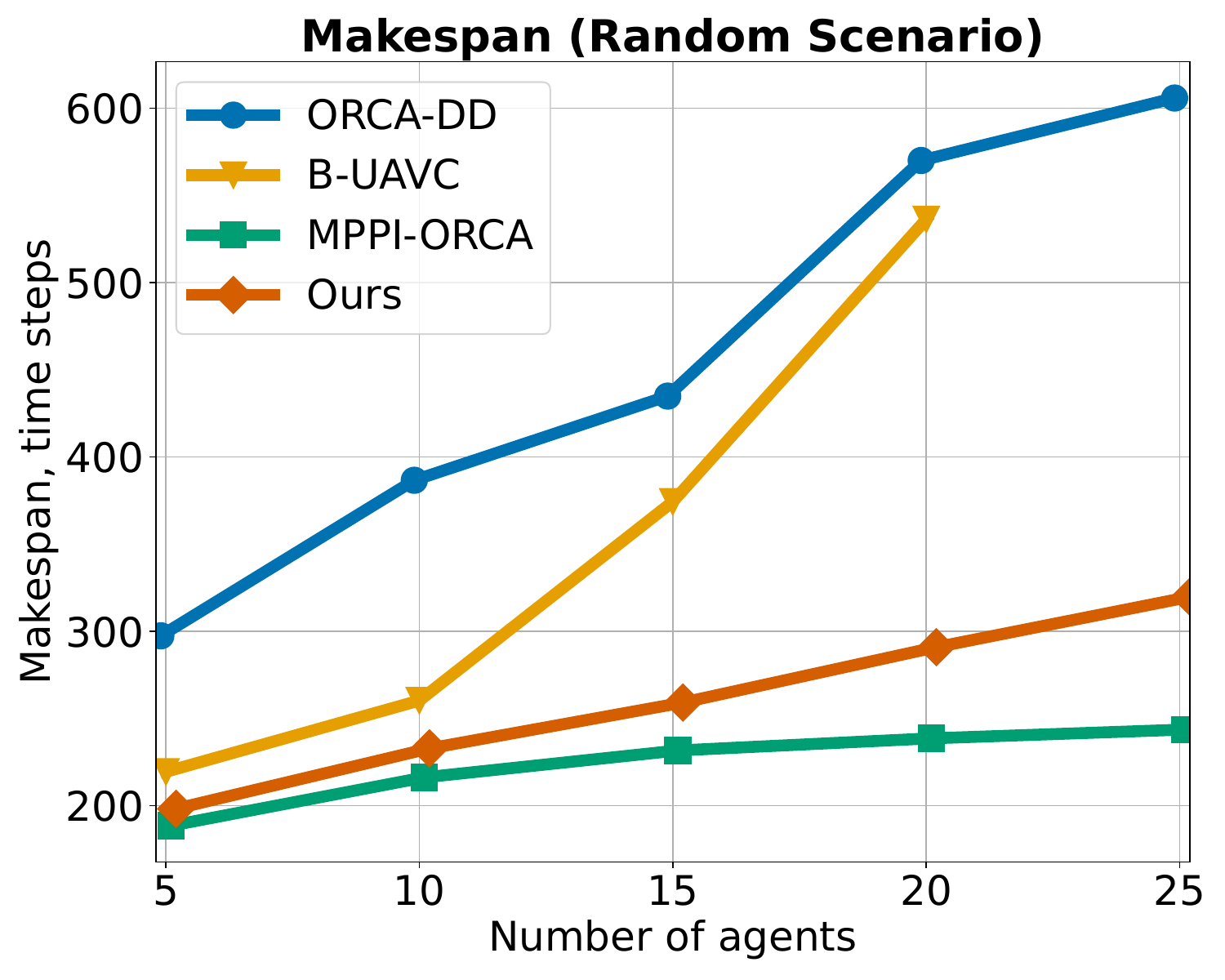}
     \end{subfigure}

    \caption{The average \textit{makespan} for evaluated algorithms for different number of agents in \texttt{Circle} and \texttt{Random} scenarios. Each point on the graph is included only if  more than 50\% of the runs were successful. The lower is the better.}%
    \label{fig:ms}
\end{figure}%

\begin{figure*}[t]
    \centering
    \includegraphics[width=0.9\textwidth]{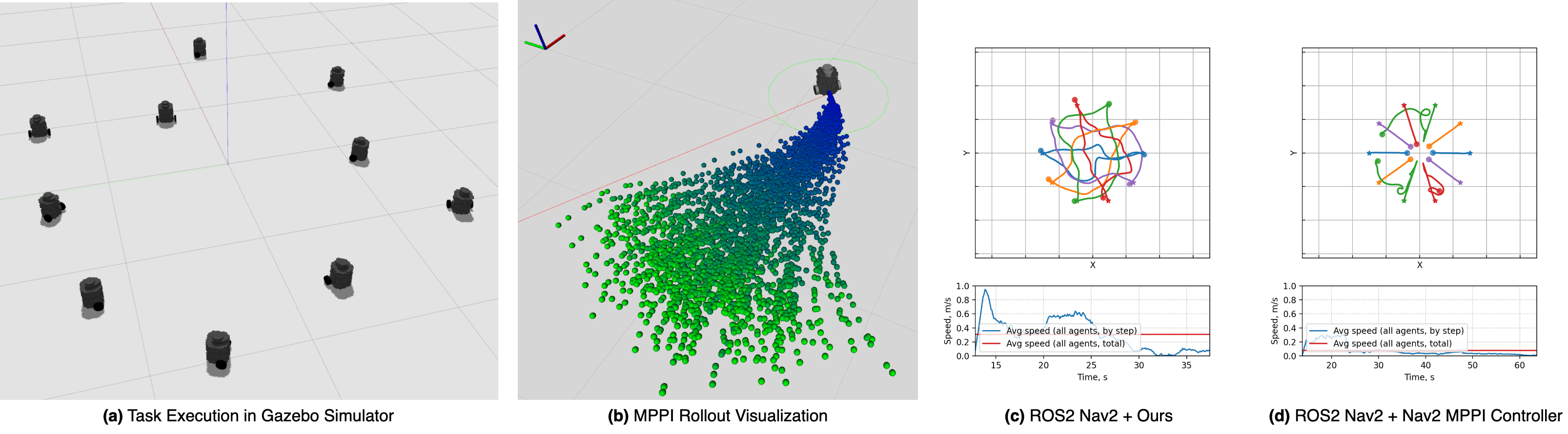}
    \caption{Simulation in Gazebo. (a) 10 Turtlebots executing the mission, (b) MPPI rollouts with sampled trajectories, (c)-(d) resultant trajectories and average velocities for the proposed approach and the built-in Nav2 MPPI Controller, respectively.}
    \label{fig:gazebo}
\end{figure*}%

We compare our method with \textsc{ORCA-DD}~\cite{snape2010smooth}, \textsc{B-UAVC}~\cite{zhu2022decentralized} and its predecessor \textsc{MPPI-ORCA}~\cite{dergachev2024model}. 

\textsc{ORCA-DD} is derived from the well-known \textsc{ORCA}~\cite{van2011reciprocal_n} algorithm but employs an increased agent radius for control computations, accounting for kinematic constraints. This method assumes that agents not only have knowledge of each other’s positions but also share internal state information, including the yaw angle and the velocity of the center of the inflated disk (which does not coincide with the center of the robot’s footprint). To more accurately simulate realistic sensing conditions, we further assumed that agents do not have access to the internal information of others, specifically the velocity of the center of the inflated disk. Furthermore, the yaw angle of other agents, used in the computations, was perturbed by noise, modeled as a zero-mean Gaussian distribution with standard deviation $\sigma_\theta = 0.1, \text{rad}$, consistent with the noise model applied to position measurements.

The \textsc{B-UAVC} method is based on the Buffered Voronoi Cells (\textsc{BVC}) approach and incorporates multiple enhancements over the base \textsc{BVC} algorithm, particularly the consideration of kinematic constraints for differential-drive robots. Unlike our approach, this method assumes the presence of noise not only in observations but also in localization. However, it presumes that agents exchange information about each other’s positions.

For ORCA-MPPI and our method, the probability of sampling a control action outside the safety constraints, denoted as  $\delta_u$, was set to $0.999$. In the proposed method, we additionally set the probabilities  $\delta_{\nu} = 0.999$  and  $\delta_{o} = 0.9975$.

\subsubsection{Experimental Results}

The primary performance indicators tracked in the experiments were the \textbf{success rate} and the \textbf{makespan}. The former is the percentage of successful runs where all agents reached their target positions without collisions. The latter is the total duration required for all agents to reach their goals. Furthermore, the presence of collisions during the execution of an instance was analyzed to assess the safety and effectiveness of the approaches.


The resultant \textbf{success rates} are reported in Table~\ref{tab:sr_comparison} as well as the percentage of runs that were terminated due to agent collisions. Evidently, the proposed approach successfully handled all tasks in both the \texttt{Circle} and \texttt{Random} scenarios, demonstrating strong safety guarantees. In contrast, its predecessor, MPPI-ORCA is highly prone to collisions. ORCA-DD and B-UAVC exhibit lower collision rates but often result in deadlocks when the robots get safely stuck and do not progress towards the goal.

Figure~\ref{fig:ms} shows the average \textbf{makespan}. Each point on the plot is included only if more than 50\% of the runs were successfully completed. Clearly, the proposed method outperforms ORCA-DD and B-UAVC in terms of solution quality, as indicated by lower makespan values. However, it performs slightly worse than the MPPI-ORCA algorithm. This difference arises from the fact that the proposed method explicitly accounts for observation and dynamic inaccuracies, leading to a more conservative motion planning strategy.

\subsection{Simulation in Gazebo}

We evaluated our algorithm using ten differential-drive TurtleBot 3 robots in the Gazebo simulation environment. The proposed method was implemented as a controller module within the ROS2 Navigation Stack~\cite{macenski2020marathon2}. For global path planning, a straight-line trajectory to the target position was employed. To simulate perception constraints, neighboring robot information was retrieved from ground truth data with added noise, following the methodology described in previous experiments. This information was then published to separate ROS topics at a frequency of 10 Hz.

The validation was conducted in a \texttt{Circle} scenario with a diameter of 6 meters. Figure~\ref{fig:gazebo} presents snapshots of the experiment along with examples of the resulting trajectories for both the proposed algorithm and the built-in MPPI controller in the Nav2 stack. The proposed approach successfully guided all robots to their target positions without collisions while maintaining real-time control selection at a frequency exceeding 10 Hz. In contrast, the built-in navigation stack algorithms were unable to effectively handle collision avoidance, highlighting the advantages of the proposed approach for multi-agent navigation tasks.

Additionally, we evaluated the Dynamic Window Approach (DWA)~\cite{fox1997dynamic}, which is also implemented within the ROS2 Navigation Stack. However, this algorithm also failed to successfully complete the task. For further details, please refer to the publicly-available supplementary video\footnote{\url{https://youtu.be/_D4zDYJ4KCk}}.

\section{CONCLUSIONS}

We have introduced a novel decentralized multi-agent collision avoidance method that integrates MPPI with a probabilistic adaptation of ORCA, addressing kinematic constraints, observation noise, and execution uncertainty. By incorporating safety-aware sampling adjustments our method improves robustness and ensures collision-free navigation. Extensive simulations demonstrated that our approach outperforms state-of-the-art methods, and validation in Gazebo confirmed its practical applicability to real-world robotic systems.






\bibliographystyle{IEEEtran}
\bibliography{IEEEabrv,IEEEexample}

\addtolength{\textheight}{-12cm}   

\end{document}